\documentclass[sigconf,screen]{acmart}

\setcopyright{cc}
\setcctype{by-nc-nd}
\acmDOI{10.1145/3757279.3785634}
\acmYear{2026}
\copyrightyear{2026}
\acmISBN{979-8-4007-2128-1/2026/03}
\acmConference[HRI '26]{Proceedings of the 21st ACM/IEEE International Conference on Human-Robot Interaction}{March 16--19, 2026}{Edinburgh, Scotland, UK}
\acmBooktitle{Proceedings of the 21st ACM/IEEE International Conference on Human-Robot Interaction (HRI '26), March 16--19, 2026, Edinburgh, Scotland, UK}
\received{2025-09-30}
\received[accepted]{2025-12-01}

\usepackage{enumitem}

\graphicspath{{figs/}}

\begin{document}

\title[Bidirectional Human-Robot Communication]{Bidirectional Human-Robot Communication for\\Physical Human-Robot Interaction}

\author{Junxiang Wang$^*$}
\affiliation{%
  \institution{Carnegie Mellon University}
  \city{}
  \country{}
}

\author{Cindy Wang}
\affiliation{%
  \institution{Carnegie Mellon University}
  \city{}
  \country{}
}

\author{Rana Soltani Zarrin}
\affiliation{%
  \institution{Honda Research Institute USA}
  \city{}
  \country{}
}

\author{Zackory Erickson}
\affiliation{%
  \institution{Carnegie Mellon University}
  \city{}
  \country{}
}
\thanks{$^*$ \href{mailto:junxiang@cmu.edu}{junxiang@cmu.edu}}

\renewcommand{\shortauthors}{Junxiang Wang, Cindy Wang, Rana Soltani Zarrin, and Zackory Erickson}


\begin{abstract}
  Effective physical human-robot interaction requires systems that are not only adaptable to user preferences but also transparent about their actions. This paper introduces BRIDGE, a system for bidirectional human-robot communication in physical assistance. Our method allows users to modify a robot's planned trajectory---position, velocity, and force---in real time using natural language. We utilize a large language model (LLM) to interpret any trajectory modifications implied by user commands in the context of the planned motion and conversation history. Importantly, our system provides verbal feedback in response to the user, either assuring any resulting changes or posing a clarifying question. We evaluated our method in a user study with 18 older adults across three assistive tasks, comparing BRIDGE to an ablation without verbal feedback and a baseline. Results show that participants successfully used the system to modify trajectories in real time. Moreover, the bidirectional feedback led to significantly higher ratings of interactivity and transparency, demonstrating that the robot's verbal response is critical for a more intuitive user experience. Videos and code can be found on our project website: \url{https://bidir-comm.github.io/}
\end{abstract}



\begin{teaserfigure}
  \includegraphics[width=\textwidth]{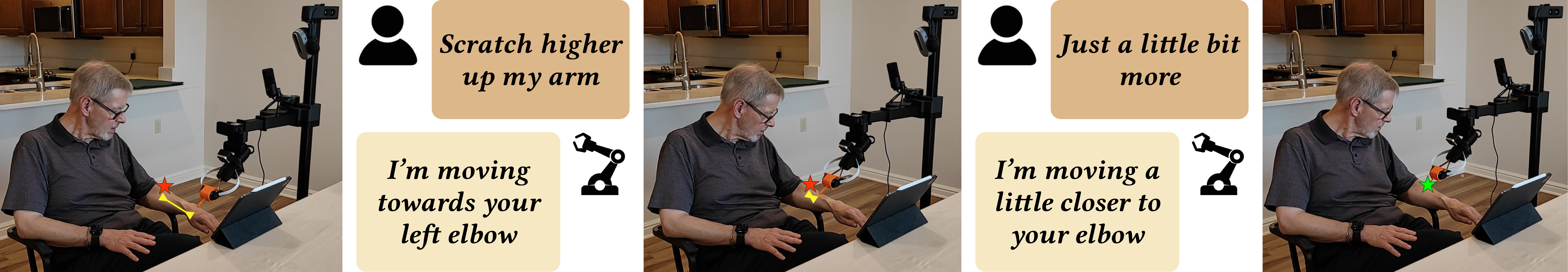}
  \caption{Example interaction with BRIDGE. User is in a scratching scenario, where they command the robot to move to the target scratching position (marked with star) through a bidirectional verbal interaction.}
  \Description{User issues two consecutive verbal commands with the second one needing interpretation only within the previous context. Each verbal command is responded with a verbal feedback from the robot. After these commands, the user is able to move the robot to the goal position.}
  \label{fig:teaser}
\end{teaserfigure}

\maketitle

\section{Introduction}

A growing body of research has shown that robots can address a range of caregiving activities in autonomous physical assistance~\cite{nanavati2023physically,erickson2018deep,gordon2024adaptable}. However, many systems lack two capabilities essential for interactive autonomy: real-time user-guided adaptation and transparent communication. Real-time adaptation allows users to adjust a robot's ongoing autonomous motion---such as tuning speed or pressure---to fit their own personal preferences and levels of comfort~\cite{jenamani2025feast,gupte2023optometrist,gopinathan2017user}. In parallel, clear communication of a robot's intent and state---such as any upcoming changes to its motions---supports higher transparency and user trust towards the robot~\cite{mehrotra2024systematic,alhaji2021trust,fischer2018increasing,wang2025cori}. 

In this paper, we propose BRIDGE, a system for \textbf{B}idirectional human-\textbf{R}obot assistive \textbf{I}nteraction with \textbf{D}ialog \textbf{G}uidanc\textbf{E}, addressing both real-time adaptation and transparency. BRIDGE allows users to issue verbal commands to change a robot's position, velocity, and force \emph{in real time} as the robot autonomously executes a planned interaction for physical assistance. To establish \emph{bidirectional communication}, our system responds to every user utterance with verbal feedback---either an \textbf{assurance} of the desired change or a \textbf{clarifying question}---thereby closing the interaction loop (an example shown in Figure~\ref{fig:teaser}).

We evaluate our system via a within-subjects user study with older adults ($n=18$) in three physically assistive tasks. Participants successfully modified the robot's position, velocity, and force through speech in real time. Beyond this capability, BRIDGE's verbal feedback to user adjustments yields higher perceived interactivity and transparency than a unidirectional ablation that allows trajectory modifications but offers no feedback, underscoring the importance of bidirectional communication.

In summary, our contributions in this paper are as follows:
\begin{itemize}[
  topsep=0pt,      
  partopsep=0pt,   
  parsep=0pt,      
  itemsep=0.1em,   
  leftmargin=1.5em,  
  labelsep=0.5em   
]
  \item We propose a bidirectional human-robot communication framework in physically assistive scenarios. This framework couples a user's trajectory commands with real-time, transparent verbal feedback from the robot, fostering a more intuitive interaction.
  \item We present a novel LLM-based pipeline that efficiently interprets user utterances in trajectory and conversation context. The pipeline simultaneously generates modifications to the trajectory's position, velocity, or force and verbal feedback through concise assurances or clarifying questions.
  \item We conduct a user study with 18 older adults and three physically assistive tasks, which demonstrates that bidirectional communication leads to higher perceived interactivity and transparency than a unidirectional ablation, while speech-based modifications remain fast enough for real-time applications.
\end{itemize}

\section{Related Works}
\label{sec:background}

\subsection{Language-guided robot motions in HRI}
The concept of influencing a robot's action through language inputs has been widely explored in different contexts, predominantly in the field of robotic manipulation. Language is often integrated in learning reward functions~\cite{liang2024learning,yang2024trajectory,sharma2022correcting}, selecting motion primitives on a high level~\cite{shi2024yell,zha2024distilling}, computing latent actions~\cite{cui2023no,karamcheti2022lila}, or training general language-guided policies~\cite{lynch2023interactive} and vision-language-action models (VLA)~\cite{kim2024openvla,intelligence2025pi}. In comparison, our method leverages the zero-shot reasoning capabilities of a general-purpose LLM. We focus on the assistive domain and utilize an LLM for real-time interpretation of utterances into trajectory modifications, without task-specific training or fine-tuning.

In physically assistive robotics, language is commonly used for issuing task-level commands, such as initiating motions in feeding assistance~\cite{padmanabha2024voicepilot,bhattacharjee2020more} or selection of predefined motion primitives~\cite{alonso2021abstraction}. Other use cases mostly revolve around commanding a robot arm for assistive object retrieval~\cite{poirier2019voice,pulikottil2018voice}. These systems are often confined to one particular assistive task, whereas our developed system can be applied to a range of physically assistive trajectories. We also make modifications to trajectories on a parameter level, directly changing all motion aspects of position, velocity, and force, without being limited to configured actions.

\subsection{Human-robot dialog systems}
While the previous section mainly focuses on voice interfaces being used only in the human-to-robot direction, many prior works also explore dialog systems between humans and robots, similar to our system. Speech serves as one of the most intuitive interfaces that can grant both personalization and transparency, especially for older adults and assistive scenarios~\cite{pradhan2020use,song2022investigation,mahmood2024situated}.

Socially assistive robots often make use of conversations in the contexts of therapy or affective support~\cite{rudzicz2015speech,lima2021conversational,spitale2023using}, while these applications generally do not involve physical interactions, which is the emphasis of our work.

Dialog can also play an important role in the domain of human-robot collaboration~\cite{wang2024mosaic,mandi2024roco,yu2025mixed,allgeuer2024robots}, and the information exchanged often revolves around task assignment, hence which party (human or robot) should be assigned with which step in a task. In contrast, we address the domain of assistive robotics, where humans can provide their preferences verbally when robots autonomously provide physical assistance. Additionally, we focus on assistive tasks that can be completed even with only human commands and no robot verbal feedback, and in this work, we look into the effect of providing such verbal feedback to humans---hence how bidirectional communication influences an interaction that may also be completed with unidirectional human-to-robot communication.

\section{Methods}
\label{sec:methods}

\begin{figure}[b]
  \centering
  \includegraphics[width=\linewidth]{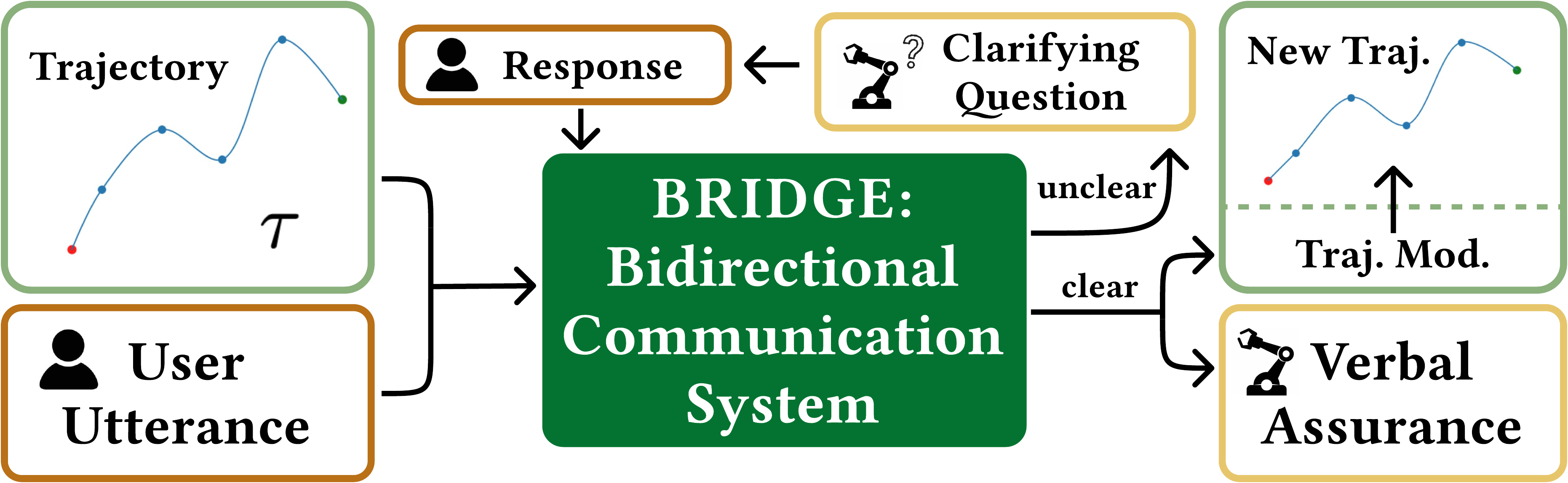}
  \caption{Flowchart of BRIDGE: our bidirectional communication system, including two cases of verbal feedback depending on whether a user utterance is clear: (1) assuring and executing any modifications to the trajectory, or (2) posing a clarification question to request for further user input.}
  \Description{The bidirectional communication system takes as inputs a trajectory and a user utterance, and either (1) infers and generates a trajectory modification from the utterance to update the trajectory, paired with verbal assurance, or (2) poses a clarifying question when the utterance is unclear, seeking additional user response.}
  \label{fig:system}
\end{figure}

Our framework takes as input (1) a planned physically assistive trajectory and (2) a user utterance, and produces real-time modifications to that trajectory based on the utterance. As outlined in Figure~\ref{fig:system}, the system either applies and communicates trajectory changes when the utterance directly implies so, or poses a clarifying question seeking for more user input. In this section, we first introduce compact representations for trajectories (\S\ref{sec:traj-rep}) and modifications to  trajectories (\S\ref{sec:mod-rep}), then discuss the LLM-based interpreter that maps an utterance to the correct modification along with a concise communication as feedback (\S\ref{sec:prompt}).

\subsection{Trajectory representation}
\label{sec:traj-rep}

\subsubsection{Assumptions}
We assume we are provided with a planned end-effector trajectory, as a sequence of 3D waypoints:
\begin{equation}
\tau=\{w_i: w_i=(t_i,\mathbf{x}_i,v_i,f_i)\}_{i=1}^N,
\end{equation}
where each waypoint $w_i$ consists of a timestamp $t_i\in\mathbb{R}_{\geq0}$, end-effector position $\mathbf{x}_i\in\mathbb{R}^{3}$, velocity magnitude $v_i\in\mathbb{R}_{\geq0}$, and desired force magnitude $f_i\in\mathbb{R}_{\geq0}$. Between consecutive waypoints, the robot end-effector is assumed to follow a straight-line Cartesian path, with velocity and force linearly interpolated in time. Force magnitudes $f_i$ are intended for assistive contact with the user and are tracked by a low-level controller implementing either impedance or admittance control. We  assume the planned trajectory interacts with a person, hence approaching certain relevant human body landmarks (major body joints such as wrists, elbows, and shoulders). We further assume access to estimated 3D positions of these landmarks, which could be obtained from body pose estimation.

\begin{figure}[tb]
  \centering
  \includegraphics[width=\linewidth]{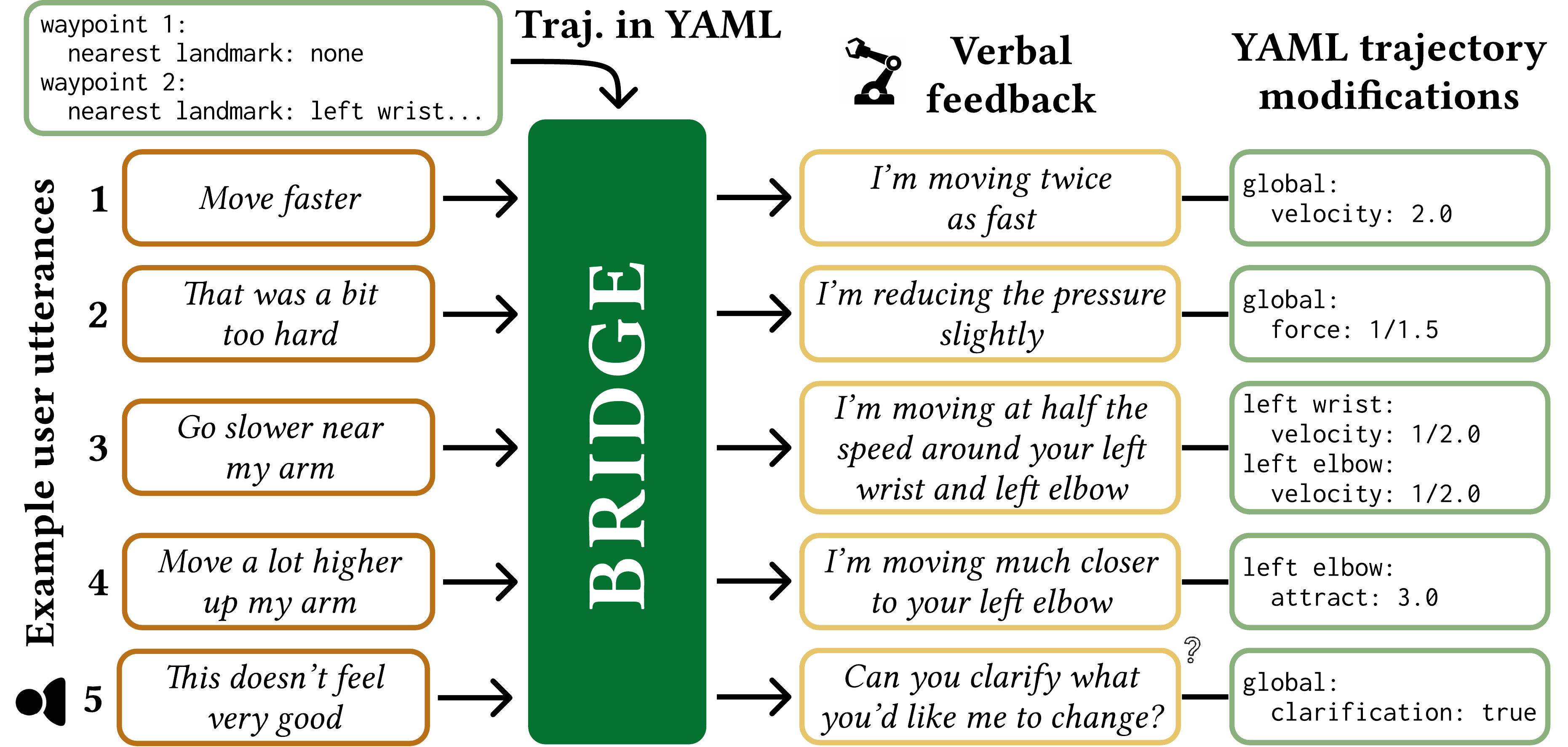}
  \caption{Example YAML trajectory and user utterances as inputs to BRIDGE, along with the generated communications and trajectory modifications in YAML format.}
  \Description{Several example user utterances with the same input trajectory, as well as our system's outputs to these examples.}
  \label{fig:examples}
\end{figure}

\subsubsection{Trajectory representation}
We build a minimal representation of the input trajectory $\tau$ that gives the LLM a high-level sketch of the planned interaction. Since the trajectory interacts with a person, we compute the nearest human-body landmark for each waypoint and use only this label to form a symbolic representation, without involving kinematic details. This representation is serialized in YAML format for structure (see top-left of Figure~\ref{fig:examples} for a snippet example). If no landmark falls within a proximity threshold, the label is left unassigned (\emph{None} in the YAML), which has no geometric implication. Such a representation supports high-level identification of where interactions occur and also provides coarse physical grounding for utterance interpretation, as we discuss in \S\ref{sec:context-interp}.

\subsection{Modification representation}
\label{sec:mod-rep}
We support trajectory modifications in position, velocity, and force, and we organize them at three scopes: \emph{global} changes that apply to the entire trajectory (\S\ref{sec:global}), \emph{landmark} changes whose influence is defined with respect to specific body landmarks (\S\ref{sec:landmark}), and \emph{waypoint} changes that affect certain waypoints (\S\ref{sec:waypoint}). These scopes do not imply priority, so if a single utterance specifies changes in multiple scopes, we apply them concurrently to the trajectory. Across utterances, changes accumulate, with velocity clamped to a fixed cap (see implementation details in \S\ref{sec:tasks-implementation}). We serialize these trajectory modifications also in YAML schema. 

\subsubsection{Global Changes}
\label{sec:global}
We support changes to the velocity and force of the entire trajectory, applied uniformly to \emph{all} waypoints. These changes correspond to broad user utterances such as ``Move faster.'' We apply these changes in a multiplicative manner, with the scaling factor represented in the corresponding YAML fields (see examples 1 and 2 in Figure~\ref{fig:examples}). Thus, a value greater than 1 indicates an increase, and a value between 0 and 1 indicates a decrease.

\subsubsection{Landmark Changes}
\label{sec:landmark}
BRIDGE also allows users to modify kinematic parameters relative to body landmarks. These changes fall into two main categories: (1) velocity and force changes around body landmarks, expressed in utterances such as ``Go slower near my arm'' (see example 3 in Figure~\ref{fig:examples}), and (2) position changes based on attraction to/repulsion from body landmarks, expressed in utterances such as ``Move up a lot higher on my arm'' (see example 4 in Figure~\ref{fig:examples}). We next elaborate on each of these categories.

\textbf{Local velocity and force changes.}\;
These changes are represented in YAML similarly to global changes, but their effect on each waypoint is subject to Gaussian decay based on the waypoint's distance from the local landmark. For example, suppose a landmark located at $\mathbf{p}_{\text{landmark}}\in\mathbb{R}^3$ has a \emph{velocity} YAML field of $k>1$, then waypoint $w$'s velocity $v$ should be increased by a factor of:\begin{equation}
  1 + (k-1)\exp\left(\frac{-\|\mathbf{p}_{\text{landmark}}-\mathbf{x}\|^2}{2\sigma^2}\right)
\end{equation}
where $\sigma$ controls the spread---a larger $\sigma$ results in wider influence around the landmark, while a smaller $\sigma$ leads to a more localized effect. We adopted the Gaussian decay model since this creates a decaying influence naturally implied by phrases like ``around my wrist.'' We set $\sigma=0.07$\,m, as empirical pilot testing shows this value creates a localized and smoothly-decaying effect. The Gaussian decay for a decrease in velocity or force is expressed similarly.

\textbf{Position changes.}\;
In order to modify the position of individual waypoints in the input trajectory, we apply the notion of artificial potential fields~\cite{khatib1986real}, commonly used in robotics for manipulation and navigation with obstacle avoidance. Attractive and repulsive potentials can therefore be placed at body landmarks, and we compute how much to displace a waypoint $w$ from the gradients of all potential functions, evaluated at the waypoint's position $\mathbf{x}$. We specify both attractive and repulsive potentials, as well as the intensity of each, via the \emph{attract} field in the YAML entry for each landmark (see example 4 in Figure~\ref{fig:examples}). Following the convention of multiplicative factors for velocity and force, a value greater than 1 indicates an attractive potential, and a value between 0 and 1 indicates a repulsive potential. For attractive potentials, we use the standard quadratic formulation. Suppose we have an \emph{attract} field of $k>1$ for a landmark located at $\mathbf{p}_{\text{landmark}}$: \begin{equation}
  U_{\text{att}}(\mathbf{x}) = \frac{k}{2}k_p\|\mathbf{p}_{\text{landmark}}-\mathbf{x}\|^2
\end{equation}
The gain $k_p$ is empirically determined to be 0.01\,m$^{-2}$. Our application differs from most manipulation and navigation scenarios in that there could be multiple attractive potentials (e.g. attraction to the forearm overall is represented as attractions to the elbow and the wrist) as opposed to a single goal. In order to ensure convergence and avoid waypoints already close to an attractive potential being pulled towards another goal, we weight each attractive potential by the inverse of its distance to the point of interest. The total attractive displacement for a waypoint located at $\mathbf{x}$ is therefore: \begin{equation}
  \Delta_{\text{att}}(\mathbf{x}) = -\sum_j \frac{w_j}{\sum_k w_k}\nabla U_{\text{att},j}(\mathbf{x}),\quad w_j = \frac{1}{\|\mathbf{p}_{\text{landmark},j}-\mathbf{x}\|}
\end{equation}
For repulsive potentials, we use the formulation with a limit distance of effect $\rho_0$. Suppose a landmark at $\mathbf{p}_{\text{landmark}}$ has $k\in(0,1)$: \begin{equation}
  U_{\text{rep}}(\mathbf{x}) = \begin{cases}
    \frac{\eta}{2k}\left(\frac{1}{\|\mathbf{p}_{\text{landmark}}-\mathbf{x}\|}-\frac{1}{\rho_0}\right)^2 \, & \text{if } \|\mathbf{p}_{\text{landmark}}-\mathbf{x}\| \leq \rho_0\\
    0 \, & \text{if } \|\mathbf{p}_{\text{landmark}}-\mathbf{x}\| > \rho_0
  \end{cases}
\end{equation}
The gain $\eta$ and the distance of effect $\rho_0$ are empirically determined to be 0.5\,m$^2$ and 0.1\,m, respectively. Therefore, assuming presence of multiple attractive and repulsive potentials, the net displacement of the position $\mathbf{x}$ of a waypoint $w$ is computed as a single step in the opposite direction of the potential field gradients: \begin{equation}
  \Delta(\mathbf{x}) = \Delta_{\text{att}}(\mathbf{x}) - \sum\nabla U_{\text{rep}}(\mathbf{x})
\end{equation}
Note that this displacement is only applied once to each waypoint, unlike the classical, iterative application of potential fields.

\subsubsection{Waypoint Changes}
\label{sec:waypoint}
Lastly, our system also supports changes to the velocity and force of individual waypoints. This capability is useful for handling utterances such as ``Go faster when you move away from me,'' which may target sections of the trajectory far away from any specific body landmark. To execute this change, the system identifies the waypoints implied by the utterance through referencing the trajectory YAML and applies the change uniformly to each one. Changes to any one waypoint does not directly affect its neighboring waypoints. Although the uniform application is similar to global changes, waypoint changes require generating a separate YAML entry for every modified waypoint.

\subsection{Interpretation of User Utterances}
\label{sec:prompt}

We design a structured LLM prompt that translates user utterances and YAML trajectories into trajectory modifications in YAML, accompanied by a brief sentence communicating back to the user. In this section, we first discuss some features that allow BRIDGE to adapt to various levels of desired change (\S\ref{sec:granularity}), as well as interpreting utterances in context of both the trajectory and the conversation history (\S\ref{sec:context-interp}). Next, we focus on the generation of robot verbal feedback, which forms the concept of bidirectional communication (\S\ref{sec:bidirectional}). Lastly, we provide some rationale and design choices regarding the compactness of LLM outputs (\S\ref{sec:compact-output}). The complete prompt used in BRIDGE can be found in the Appendix.

\subsubsection{Granularity}
\label{sec:granularity}
The prompt is designed to handle both generic and fine-grained adjustments to the motion aspects of position, velocity, and force. For a generic adjustment without specifying further granularity (e.g. ``Go slower''), we configure the change magnitude $k$ to a default factor of 2 ($k=2$ for increases and $k=0.5$ for decreases), as we find this to be an empirically distinguishable magnitude for typical assistive trajectories. When the user desires more fine-grained control, (e.g. ``Go slightly slower'' and ``Press much harder''), the prompt provides examples that allow the LLM to reason about the implied magnitude from the utterance; see examples 2 and 4 in Figure~\ref{fig:examples} for sample utterances and their corresponding YAML modifications. Finally, we bound the maximum change magnitude to a factor of 3 to avoid drastic changes. 

\subsubsection{Context-aware Utterance Interpretation}
\label{sec:context-interp}

To create a fluid and intuitive interaction, the prompt contains two sources of context---planned trajectory and conversation history---enabling the LLM to interpret ambiguous commands without requiring excessive specificity from the user.

\textbf{Trajectory context.}\;
The prompt's trajectory context is provided via the YAML representation introduced in \S\ref{sec:traj-rep}, which specifies the nearest body landmark for each waypoint. This context allows the LLM to resolve spatial ambiguities. For instance, a user does not need to specify whether they mean their ``left'' or ``right'' elbow, as this information can be inferred from the planned motion. Trajectory context also enables correct understanding of high-level references, like a command relating to an entire ``arm.'' Even though ``arm'' is not a specific body joint, the LLM can use trajectory information to deduce which landmarks (e.g., shoulder, elbow, wrist) are relevant to the user's command (see examples 3 and 4 in Figure~\ref{fig:examples}).

\textbf{Conversation context.}\;
The prompt also incorporates context from conversation history, specifically the most recent verbal exchange (user utterance, YAML trajectory changes, and robot verbal feedback). This context allows the LLM to correctly interpret follow-up commands. For example, the user may say ``Go faster'' and a subsequent command of ``A little bit more.'' While vague in isolation, this second utterance can be correctly interpreted by the LLM as another, smaller velocity increase. Retaining conversation history also enables reversing changes with utterances such as ``Undo that'' or ``Forget what I just said.'' This conversational memory allows users to make iterative refinements naturally, as shown in Figure~\ref{fig:teaser}.

\subsubsection{Bidirectional Communication}
\label{sec:bidirectional}
Bidirectional communication, a key feature of BRIDGE, is achieved by providing verbal feedback for each user utterance. We expect that the format of dialog in general, regardless of the exact content, will enhance the perception of interactivity, which is important for physically assistive scenarios. Specifically, BRIDGE provides two different types of feedback to handle different utterances: offering assurance for any utterances that imply changes to the robot's trajectory, or proactively asking a clarifying question when the meaning of an utterance is unclear. The aim of both types of feedback is to fully communicate the robot's internal state to the user to support mutual understanding.

\textbf{Assurance for change-making utterances.}\;
When a desired modification can be extracted from a user utterance, the LLM also generates one concise sentence to assure the user of the upcoming motion modification (see examples 1--4 in Figure~\ref{fig:examples}). These assurances are generated without involving technical details, only communicating the magnitude of change when it is easy to interpret from a user perspective (see examples 1 and 3 in Figure~\ref{fig:examples}).

\begin{figure*}[tb]
  \centering
  \includegraphics[width=\linewidth]{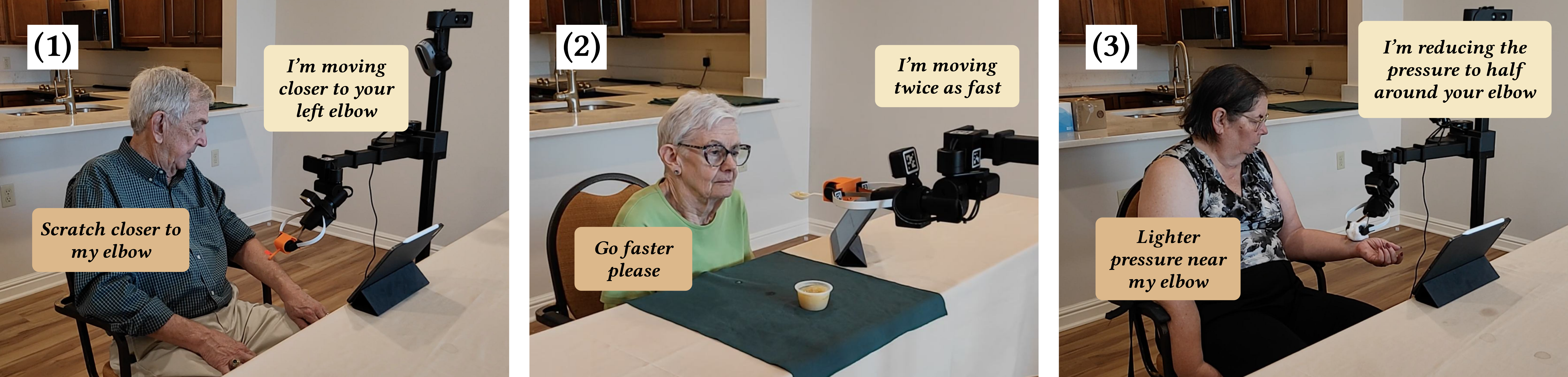}
  \caption{Snapshots from all three tasks implemented for the user study ((1) scratching, (2) feeding, and (3) bathing), with example user utterances (orange) and verbal responses from the robot (yellow) generated by BRIDGE.}
  \Description{One image per task showing task details and example verbal exchanges between the user and the robot. In the scratching task, the robot's scratching tool is in contact with the user's left arm. In the feeding task, the robot is holding a spoonful of applesauce and going towards the user. In the bathing task, the robot is holding a piece of washcloth, in contact with the user's left wrist.}
  \label{fig:tasks}
\end{figure*}

\textbf{Clarifying questions for unclear utterances.}\;
When an utterance is ambiguous or does not map to an adjustable parameter, even considering trajectory and conversational context, BRIDGE seeks clarification from the user. Such utterances could be a general expression of feeling (e.g., ``This doesn't feel good''), or it could come from incorrect speech detections, which are common in real-world environments. In these cases, the system makes no modification to the trajectory and instead produces a clarifying question to ask for more information.
These questions are deliberately kept open, rather than suggesting specific options, to ensure that users retain control over expressing adjustments. Upon the user's response, a second-stage prompt is then constructed to query the LLM for a new YAML block based on the user's clarification. This prompt is designed to be much more concise than the main prompt and contains only the immediate context of the question and answer, to facilitate minimal response latency from the LLM. This process of posing clarifying questions is iterative, until the user's intent is no longer ambiguous (as shown in Figure~\ref{fig:system}). We use a flag in the YAML representation of trajectory modification to indicate whether further clarification is required (see example 5 in Figure~\ref{fig:examples}). These clarifying questions are phrased in plain conversational language, hence easy for users to interpret and act on. 

\subsubsection{Compactness of Response}
\label{sec:compact-output}
Because LLMs generate responses autoregressively---one token at a time---the total response latency is directly influenced by the number of output tokens. To minimize latency and maximize response speed, we design the prompt to require the shortest possible response. Specifically, the output is restricted to only the YAML fields that contain modifications (change magnitude $k\neq 1$). Utterances unrelated to trajectory edits (e.g., ``I'm happy today'') are also ignored entirely.
Most importantly, we prioritize \emph{landmark} changes over \emph{waypoint} changes. A command like ``Move faster near my wrist'' could be represented by modifying a list of individual waypoints, but this would result in a verbose output with separate entries for each waypoint. Instead, we prioritize using a landmark change, which not only provides a more compact representation but also offers the smoother exponential decay described in \S\ref{sec:landmark}. Collectively, these measures ensure the LLM's output is concise, minimizing latency and creating a more interactive experience.

\section{User Study}
\label{sec:study}
We conducted a \emph{within-subjects} user study to evaluate BRIDGE's efficacy for real-time trajectory modification and to measure the specific contribution of bidirectional verbal feedback. Specifically, we test the following two hypotheses:

\textbf{H1 (Efficacy)} -- Users will be able to use BRIDGE to effectively modify the position, velocity, and force of planned trajectories for different physically assistive tasks.

\textbf{H2 (Contribution of Bidirectional Feedback)} -- Bidirectional verbal feedback from the robot will facilitate a more interactive and transparent experience for users, compared to a no-feedback method with the same ability to make trajectory changes.

\subsection{Tasks and Implementation}
\label{sec:tasks-implementation}
We designed three different assistive tasks, where each task focuses on a different aspect of the motion (position, velocity, or force) for the user to make modifications to. All tasks were performed autonomously with a Stretch 3 robot, a mobile manipulator with a 5-DoF arm and a gripper.
\begin{enumerate}[
  topsep=4pt,      
  partopsep=0pt,   
  parsep=0pt,      
  itemsep=0.1em,   
  leftmargin=2em,  
  labelsep=0.5em   
]
  \item \textbf{Scratching (position)}: The robot held a 3D-printed scratching tool and began scratching near the participant's left wrist. Participants were given the goal of modifying the position of scratching to an area on the upper forearm, indicated by stickers placed on the participant's arm.
  \item \textbf{Feeding (velocity)}: The robot held a spoon and scooped from a bowl of applesauce to feed the participant a total of three times. Participants were given the goal of modifying the velocity of feeding to a level they were comfortable with. To encourage participants to actively adjust speed, the initial motion was intentionally designed with a slow velocity.
  \item \textbf{Bathing (force)}: The robot held a piece of dry washcloth and wiped the participant's left forearm from the wrist to the elbow, a total of four times. Participants were given the goal of modifying the force of bathing to their liking. Given the subjective nature of force preference, the primary purpose of this task was to ensure participants felt empowered to make adjustments, rather than to converge on a specific target force.
\end{enumerate}
Figure~\ref{fig:tasks} shows one snapshot from each task during the user study, along with sample user utterances and the corresponding response from the robot. Despite each task focusing on one motion aspect, participants were welcomed to change more than one aspect in each task, or even in the same utterance. We used Microsoft Azure's speech-to-text service to transcribe user speech. Once the user finished speaking, the service sent the complete utterance to the robot, which then paused its motion and queried the LLM (GPT-4.1) to generate desired modifications. The query would take approximately 1-2 seconds, an interval kept short by our compact YAML representation. When the query finished, the robot updated its trajectory and then restarted its motion along the new trajectory.

Multiple utterances in the same interaction were treated as cumulative, so e.g. the second utterance would act on the modified trajectory from the first utterance. The maximum velocity of the robot end-effector was bound to 0.1\,m/s for safety.

\begin{figure*}[tb]
  \centering
  \includegraphics[width=0.9\linewidth]{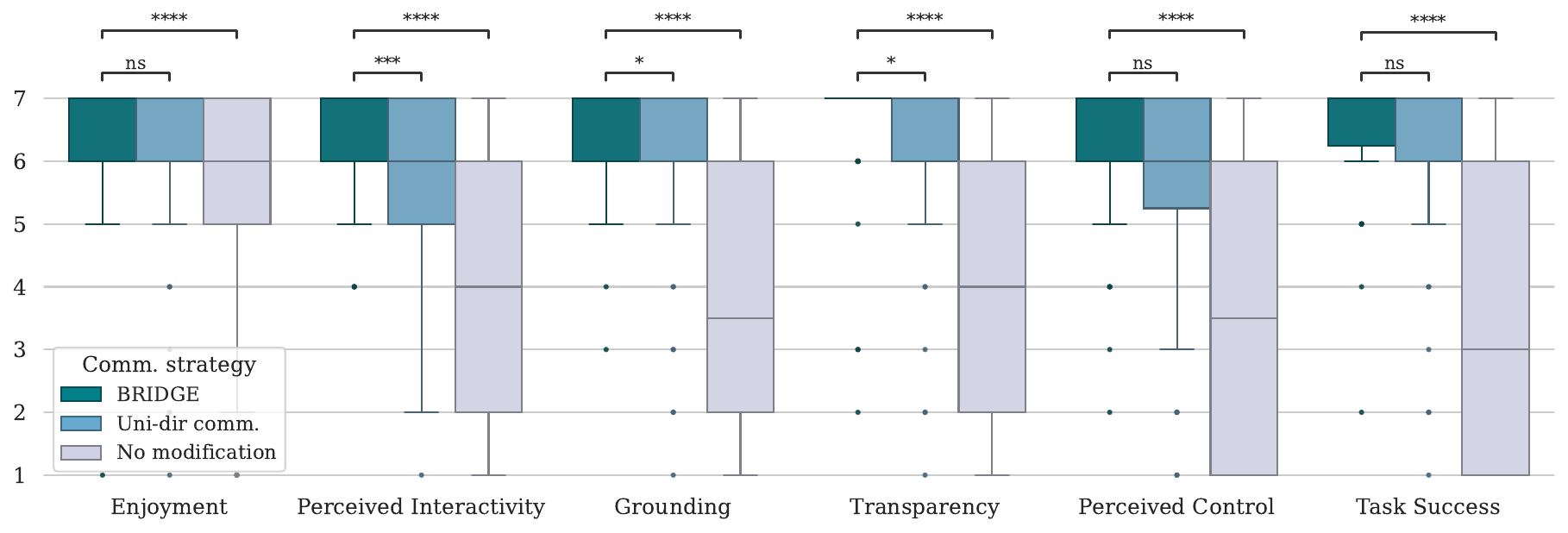}
  \caption{Box plots showing the distribution of survey responses across all participants and tasks. After fitting ordinal mixed-effects models to each question, we conduct Wald tests to assess pairwise differences between BRIDGE and both the no communication baseline and the unidirectional communication ablation. ``ns'' denote lack of significant difference, and asterisks denote significance levels ($p<0.05$, $p<0.01$, $p<0.001$, $p<0.0001$).}
  \label{fig:box-plot}
  \Description{Our method is rated highly and outperforms the no-modification baseline with significant difference on all six Likert items. Compared with the unidirectional ablation, our method is rated significantly higher in perceived interactivity.}
\end{figure*}

\subsection{Participants and Setting} 
We recruited $n=18$ older adults from a local independent living community (7 male/11 female; age range 74--90 with $M=82.1$ and $SD=4.4$). Only two participants reported any experience with autonomous robots in general (levels 2 and 3 on a 5-point scale; all other participants reported no experience).

The study was conducted in an empty apartment within the community, with all tasks completed in a single session. The study design, the experiment protocol, and the consent forms received approval from our Institutional Review Board.

\subsection{Communication Strategies and Procedure}
To test our hypotheses, we designed three communication strategies that participants experienced in a within-subjects manner:
\begin{itemize}[
  topsep=4pt,      
  partopsep=0pt,   
  parsep=0pt,      
  itemsep=0.1em,   
  leftmargin=1.5em,  
  labelsep=0.5em   
]
    \item \textbf{BRIDGE}: This is our full proposed bidirectional communication system. The robot would modify its trajectory according to the user's verbal command and provide verbal feedback in the form of either an assurance or a clarifying question.
    \item \textbf{Unidirectional communication} (Ablation): This strategy is designed to isolate the effect of the robot's verbal feedback. The robot can still modify its trajectory based on user commands, but it provides no verbal assurance or clarifying questions. If an utterance is unclear, the robot would simply pause and then resume its previous motion without change.
    \item \textbf{No-modification strategy} (Baseline): In this strategy, the robot would still listen for user speech and pause its motion, but it would not make any modifications to its trajectory. After the standard pause, it would always resume its original, unmodified path. This baseline is designed to measure the efficacy of trajectory changes in our method and the ablation.
\end{itemize}

Each participant completed nine interactions with the robot. Each interaction (trial) consisted of performing one of the three assistive tasks (scratching, feeding, or bathing) with one of the three communication strategies above. A participant would complete all three trials for a single task---hence experiencing all communication strategies for the task---before proceeding to the next task. To mitigate ordering effects, we counterbalanced the sequence in which the three tasks were presented to each participant, as well as the order of the three communication strategies within each task.

\subsection{Measures}
After each trial, the participants were asked to answer a survey with the following Likert items on a 7-point scale (7 for strongly agree, 1 for strongly disagree):
\begin{enumerate}[
    topsep=4pt,      
    partopsep=0pt,   
    parsep=0pt,      
    itemsep=0em,   
    leftmargin=2em,  
    labelsep=0.25em,   
    label=L\arabic*.
]
    \item \emph{(Enjoyment)} I enjoyed the interaction.
    \item \emph{(Perceived Interactivity)} The robot felt interactive and responsive.
    \item \emph{(Grounding)} I was confident the robot understood what I meant.
    \item \emph{(Transparency)} I could tell exactly what changed in the robot's motion after my input.
    \item \emph{(Perceived Control)} I felt in control of the robot's motions.
    \item \emph{(Task success)} At the end of the trial, I was able to achieve the overall task objective.
\end{enumerate}
The italicized terms inside parentheses denote the high-level concepts each item was designed to evaluate and were not shown to the participants.

We also logged the following data for quantitative analysis: the content of each user utterance, the LLM's full response (YAML and verbal feedback) and processing latency (i.e., the motion pause duration), and the interaction timestamp for each utterance.

\section{Results and Discussion}
\label{sec:results}

Figure~\ref{fig:box-plot} shows the distribution of all participants' responses to the Likert-item questions. We fit ordinal mixed-effects (proportional-odds) models with random intercepts for participant, task, and their interaction effects. Communication strategy was held as a fixed effect. Omnibus effects were assessed with a likelihood-ratio test. Pairwise differences between our method and the baseline or the ablation were evaluated with Wald tests from the fitted model, with Holm adjustment for multiple comparisons.

\begin{figure}[tb]
  \centering
  \includegraphics[width=\linewidth]{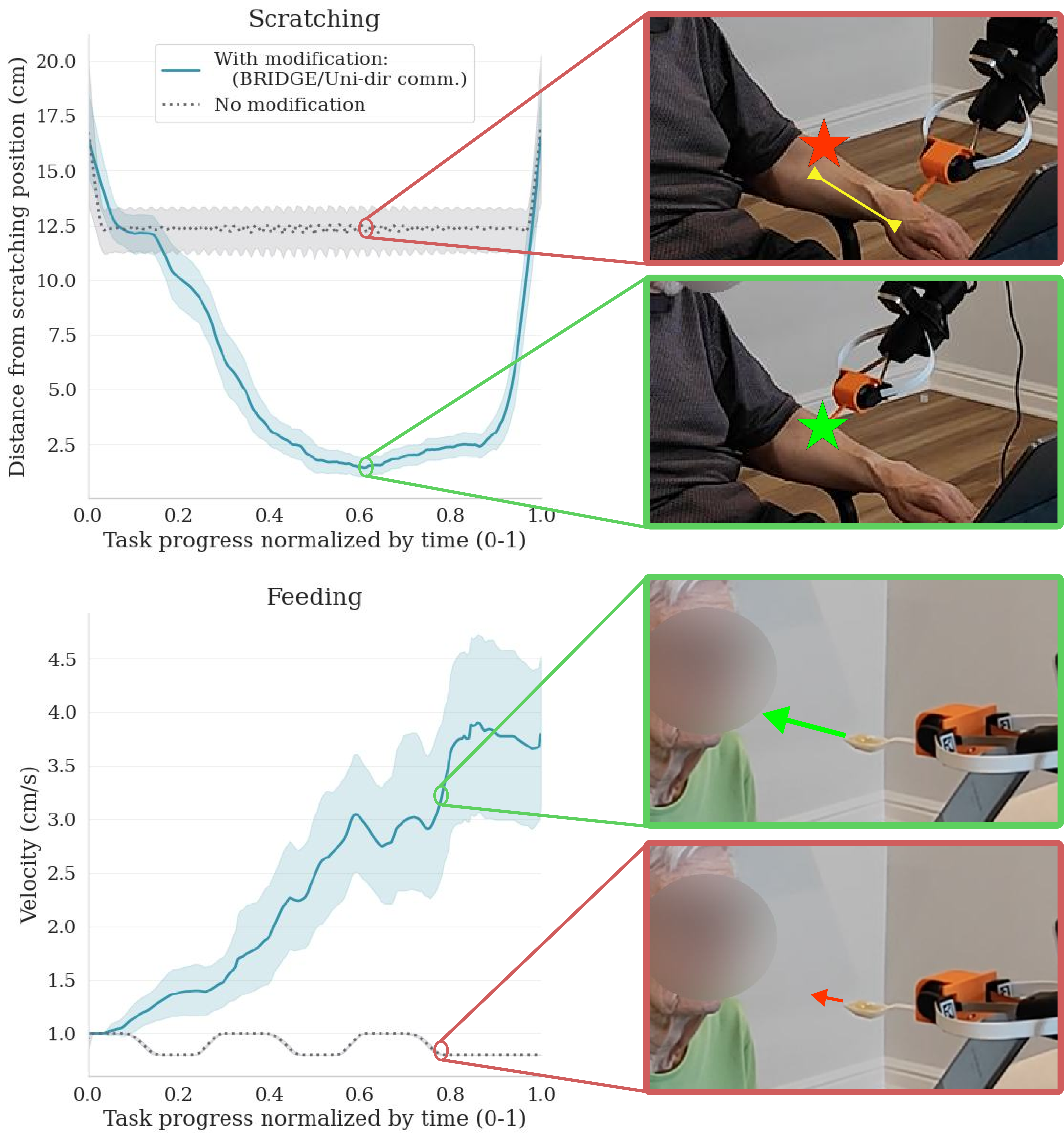}
  \caption{Efficacy of trajectory modifications, averaged over all participants. The plots show changes in position for the scratching task (top) and velocity for the feeding task (bottom) over normalized task progression. They compare the communication strategies that allow modifications (BRIDGE and the unidirectional ablation, solid lines) against the no-modification baseline (dotted line). The snapshots on the right visualize the difference in robot state, comparing a trial with modifications to one without at a representative point in the interaction.}
  \Description{Curves for strategies with modification and the no-modification baseline clearly diverge for both parameters plotted. Pictures on the right visualize closer proximity to target in the scratching task and higher velocity in the feeding task.}
  \label{fig:change}
\end{figure}

Figure~\ref{fig:change} provides objective evidence for our efficacy hypothesis (H1), illustrating how users successfully modified the robot's motion to achieve the task goals. As per our task design, we visualize the two tasks that were designed with clear, objective targets. The plots show the robot's state over the normalized task progression: the top plot shows the robot's proximity to the target position for the scratching task, while the bottom plot shows the robot's velocity for the feeding task, both averaged across all participants. We do not visualize the results for the bathing task since controlling the force is guided by subjective preference rather than a desired target. In the visualized tasks, the successful modifications are evident in the clear divergence of the trajectories for the modifiable strategies (BRIDGE and the unidirectional ablation, solid lines) compared to the static, preplanned path of the no-modification baseline (dotted line). Variations in the baseline reflect the designed dynamics of the original trajectory. To complement the plotted data, the snapshots on the right of Figure~\ref{fig:change} visualize the effect of user commands, where colored stars and arrows (red for without modification, green for with modification) visually depict the difference in robot state at a representative point in the interaction.

\subsection{Modification Efficacy}
\label{sec:efficacy}

As shown in the top plot for scratching in Figure~\ref{fig:change}, users were able to command the robot to scratch at the target location under the bidirectional and unidirectional strategies (as proximity goes towards zero). In contrast, under the no-modification baseline, the robot's motion was unaffected by user commands and simply followed its original, pre-planned trajectory. The accompanying snapshots on the top-right of Figure~\ref{fig:change} provide a visual confirmation,  showing the robot successfully reaching the target area (marked with star) in trial with modifications. Similarly, the bottom plot for feeding in Figure~\ref{fig:change} shows that users were able to command substantial increases in the robot's speed under the bidirectional and unidirectional strategies, increasing it by a factor of three to four compared to the cautious initial trajectory. In contrast, the velocity profile in the baseline method remained unchanged and followed the preplanned trajectory. This demonstrates the effectiveness of our underlying speech-to-modification system for changing different aspects of a robot's trajectory.

To complement the objective data, we analyzed participants' ratings of efficacy to the Likert items as shown in Figure~\ref{fig:box-plot}. Participants reported a high degree of task success (L6) and perceived control (L5) when using BRIDGE or the unidirectional ablation. Statistical tests revealed that both strategies were rated significantly higher than the baseline for both L6 ($p<0.0001$) and L5 ($p<0.0001$). Importantly, the high task success ratings hold across all three tasks, confirming the effectiveness of our method for modifying the intended aspect of each task: position for scratching, velocity for feeding, and force for bathing (see Figure~\ref{fig:task-success}).

Moreover, there was no significant difference in task success or perceived control between our bidirectional method and the unidirectional ablation. This finding confirms that both strategies were equally effective at achieving task goals of modifying trajectories, which carries the crucial implication that any differences observed in other subjective ratings (discussed in \S\ref{sec:perception}) can be attributed directly to the presence of the robot's verbal feedback.

Finally, the system's efficiency was confirmed by its low latency. The average time from a user finishing an utterance to the robot executing the modification was measured to be 1.7\,s across all trials with modifications, which consists of 1.3\,s for the LLM query and 0.4\,s for the motion planner. This rapid response time directly results from our emphasis on compact representations in LLM outputs and confirms the viability of our system for real-time interaction.

In summary, these three sources of evidence---the objective success in modifying trajectories, the high ratings of task success and user perceived control, and the low system latency---collectively support our \textbf{efficacy hypothesis (H1)}. The results confirm that our system is effective at performing real-time trajectory modifications.


\subsection{User Perception and Feedback}
\label{sec:perception}

To test H2, we analyzed the participants' perception of their interactions as reported in survey ratings (shown in Figure~\ref{fig:box-plot}). BRIDGE was rated significantly higher than the no-modification baseline across all measures ($p<0.0001$).
More importantly, when compared to the unidirectional ablation, BRIDGE was perceived as significantly more interactive (L2, $p<0.001$). Participants also expressed higher confidence that the robot understood them (L3, $p<0.05$) and found it easier to discern changes in the robot's motion (L4, $p<0.05$).

This quantitative preference is supported by qualitative feedback gathered in post-study debriefings. When asked which communication strategy they preferred, all participants selected the bidirectional communication by BRIDGE and provided insight into the two types of verbal responses: assurances and clarifying questions.

Verbal assurances elicited varied preferences among the participants. Many appreciated its value: P15 noting that with assurances, ``you know the robot has interpreted what you want.'' P18 felt that assurances made them ``prepared for what's going to happen.'' However, some participants took a more indifferent stance: P5 said ``It doesn’t matter'' whether the assurances were communicated, and P8 stated that ``[the robot] didn’t need to repeat to [them] what it’s going to do.'' P13 offered the nuance that while ``reassuring'' initially, assurances could become ``annoying'' in long-term use.

In contrast, participants were universally positive about the merit of clarifying questions. P5 said ``If [the robot] doesn’t understand, then [they] want to know,'' and P15 noted that without clarifying questions, they ``don’t know if [they] said something that the robot wasn’t understanding.'' P13 stated that clarifying questions provide a method for them to ``learn from [the robot].'' Throughout the user study, BRIDGE generated a clarifying question in response to 17\% of all user utterances, demonstrating its role as a key mechanism for resolving ambiguity.

Overall, this varied reception to different types of feedback suggests an opportunity for personalization, such as tuning the frequency or verbosity of verbal feedback to user preference and adapting over time with familiarity. 

In summary, through both the survey analysis and qualitative feedback, results show that BRIDGE was strongly preferred over the unidirectional ablation, fostering a more interactive and transparent experience, hence supporting \textbf{H2}.

\subsection{Perceptual Bias and the Need for Transparency}

\begin{figure}[tb]
  \centering
  \includegraphics[width=\linewidth]{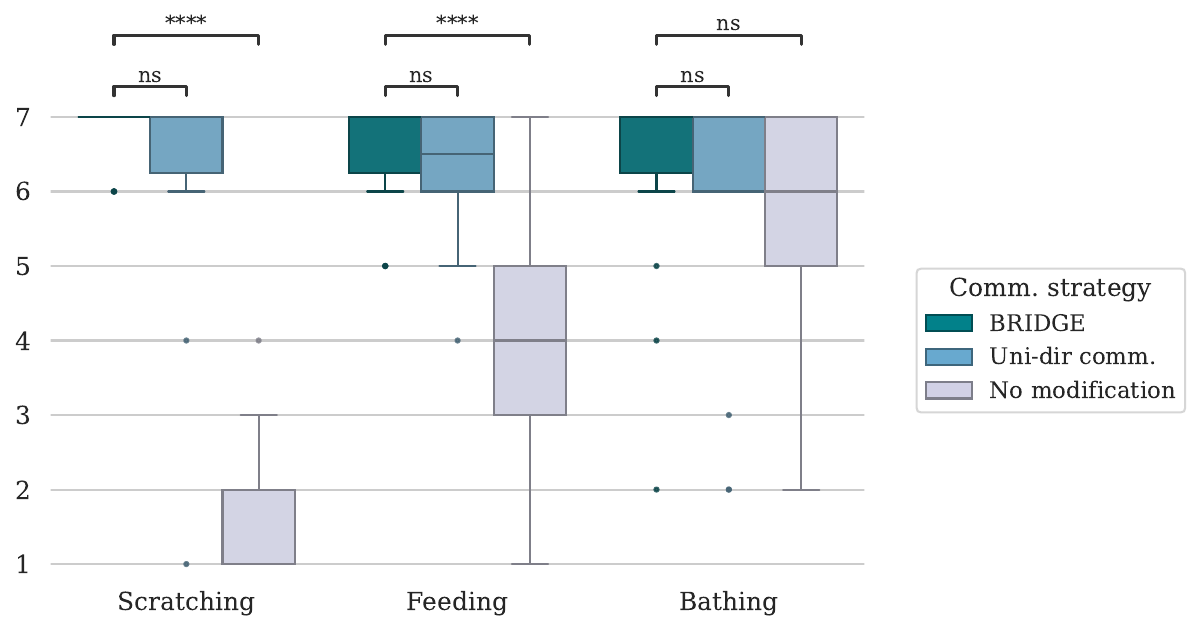}
  \caption{Box plot showing the distribution of survey responses for perceived task success (L6), separate for each task. ``ns'' denote lack of significant difference, and asterisks denote significance levels ($*$$*$$*$$*$ means $p<0.0001$).}
  \Description{Our bidirectional method and the unidirectional ablation both receive high task success ratings for all tasks, but for the no-modification baseline, it is rated the highest and comparable to the other strategies in bathing, rated less in feeding, and least in scratching. Significant differences are observed for both scratching and feeding, but not for bathing.}
  \label{fig:task-success}
\end{figure}

An interesting finding emerged when we analyzed the perceived task success (L6) in a per-task manner, as shown in Figure~\ref{fig:task-success}. The data reveals a stark contrast between tasks with objective versus subjective goals.

The scratching task had a clear, objective target, and participants correctly identified the baseline's failure to react to their inputs, rating its success very low (upper quartile of 2). This was significantly different from our method, which was rated very successful.

However, for the bathing task, where the ``correct'' force is subjective, participants reported a high degree of success (lower quartile of 5) even for the baseline method, with no statistically significant difference from our method. Such ratings suggest a powerful perceptual bias: when a change is difficult to perceive and driven by user command, participants tend to believe their command was successful, even when the robot's behavior remained the same. This interpretation is supported by qualitative feedback, where participants claimed they perceived commanded force changes, and sometimes velocity changes in the feeding task too, even when experiencing the baseline method. Our observation finds psychological grounding in research on causal attribution~\cite{kelley1980attribution} and the illusion of control~\cite{langer1975illusion}, where people may infer causal relationship between their own action and a temporally subsequent event.

This finding underscores the importance of transparency in physical human-robot interactions. When a user's perception can diverge from the robot's actual behavior, the liability is on the robot to provide clear and transparent feedback. Without such grounding, the user may develop an inaccurate mental model of the system's capabilities, leading to frustration and mistrust in the long run.

This result further justifies the need for clarifying questions as well. Between BRIDGE and the unidirectional ablation, a major functional difference is that utterances that trigger a clarifying question in BRIDGE will not change the robot's motion in the ablation. Despite the two both having high perceived success, the ablation places the burden on users to notice and correct errors---especially difficult given the perceptual bias we observe. With clarifying questions, such uncertainty surfaces immediately, and users enjoy a more intuitive experience, as reflected by qualitative comments.


\section{Conclusion}

We present BRIDGE, a framework for bidirectional human-robot communication during physically assistive scenarios, where users can verbally modify a robot's trajectory in real time across the motion aspects of position, velocity, and force. Our method leverages an LLM to efficiently translate any user utterance into compactly represented trajectory modifications while simultaneously generating appropriate verbal feedback---either as an assurance of a desired change or as a clarifying question. A user study with 18 older adults demonstrates the efficacy of our method for trajectory modifications and proves the need for the robot's bidirectional verbal feedback, which significantly enhances user experience by improving perceived interactivity and transparency.

\textbf{Future work.}\;
Our work has a few limitations that could open avenues for future research. First, our user study only involved a single session per participant, so how to appropriately structure bidirectional verbal response in a long term deployment scenario remains unexplored. Additionally, our system can modify a range of physically assistive planned trajectories assuming the user stays stationary, but how to effectively adapt BRIDGE to real-time policies or general manipulation settings remain open questions. Moreover, testing BRIDGE on more dynamic tasks and scenarios may offer additional insight into the value of each component of our system. Lastly, spoken language may not be the most intuitive form of communication at all times, and other methods of communication (e.g. tactile, gestures) could be considered as well.

\begin{acks}
This work is supported by Honda Research Institute USA.
\end{acks}

\bibliographystyle{ACM-Reference-Format}
\bibliography{references}

\clearpage
\appendix

\section{Full LLM Prompt}
Most of the LLM prompt we used is fixed, but there are sections towards the end that we dynamically populate based on the trajectory, user utterance, as well as conversation history. We first present the fixed content in the prompt, where we mark the areas to be dynamically populated with bold italicized font, and we give some examples for those dynamic contents afterwards.
\subsection{Fixed content}
{\setlength{\parindent}{1em}
You are a robot with a planned trajectory defined by a sequence of waypoints to move to for interacting with what you see around you. Each waypoint specifies the position, velocity, and force (pressure) of your gripper that interacts with a person. You will be given:\\
- A YAML dictionary of local waypoints with their nearest body landmark.\\
- A list of all body landmarks that you detected.\\
- The most recent sequence of user utterances and your own prior YAML responses, if they exist.\\
- The user's current utterance made while you are moving.\\
\\
Based on the user's current utterance, you should output a YAML block in the below format
\begin{verbatim}
  yaml
waypoint [x]:
    force: [multiplier]
    velocity: [multiplier]
...
global:
    clarification: true | false
    force:[multiplier]
    stop: true | false
    velocity: [multiplier]
[body landmark]:
    attract: [multiplier]
    force: [multiplier]
    velocity: [multiplier]
...
\end{verbatim}
Where:\\
- Each numeric field represents a multiplier relative to the current value:\\
\indent - All values are always > 0\\
\indent - Values greater than 1.0 mean the person wants that quantity to increase (e.g., ``faster'', ``firmer'', ``push harder'') or attract the robot.\\
\indent - Values between 0.0 and 1.0, always represented as a fraction, mean the person wants that quantity to decrease (e.g., ``slower'', ``gentler'', ``less force'') or repel the robot. These values should be represented as a fraction with a numerator of 1 (e.g., 1/2.0).\\
\indent - 1.0 means no change.\\
- `\verb|stop: true|' means the user has requested that the robot immediately stop moving.\\
- `\verb|clarification: true|' means the robot requires a follow-up clarification.\\
For the attract, velocity, and force fields, if the request's language implies gradation (e.g., ``just a bit faster'' or ``way too rough''), adjust the magnitude accordingly:\\
- Default change in intensity is a multiplier of 2.0 (double) for increasing change or 1/2.0 (half) for decreasing change.\\
- The max increase should be around 3.0, and max decrease around 1/3.0.\\
- Infer how strong or subtle the change is from modifiers like ``a little'', ``slightly'', ``a lot'', ``way too'', ``much more'', etc. and choose an appropriate multiplier based on this reasoning knowing that the default is to double of half. Use your own judgment to map qualitative descriptions into meaningful quantitative adjustments.\\
The following is a list of all detected body landmarks: left wrist, right wrist, left elbow, right elbow, left shoulder, right shoulder, mouth\\

Adjustments should be categorized into one of three types:\\
1. Body Landmarks: When a request references a specific body landmark, update only its corresponding entry or entires included within that body landmark (e.g. ``foot'' and ``knee'' for ``leg''). Body landmark entries also include an ``\verb|attract|'' field to reflect movement preferences, where:\\
\indent - attract > 1.0: move closer to the body landmark (e.g., ``stay closer to my left side'')\\
\indent - attract < 1.0: move farther away (repel) (e.g., ``stay away from my knee'')\\
\indent - attract = 1.0: no change.\\
2. Global: When a request affects the overall trajectory (e.g. ``go faster'', ``use less pressure'', ``finish this task slower'').\\
3. Local Waypoints: When a request targets a specific section of the trajectory without mentioning a landmark (e.g., ``go slower on the way toward me'').\\
Rules for applying changes:\\
- Body landmark references: Only modify the landmark entry. Assume that the body landmarks listed in the given YAML are the only relevant entries.\\
- Local vs. Global: Default to global unless the request clearly refers to a specific part of the trajectory.\\
- Multiple changes: Multiple values can be modified by one request, but do not apply both local and global changes for the same quantity unless the request clearly calls for both.\\
- Stop parameter: only set 'stop' to true if the user explicitly says ``stop,'' and treat phrases like ``stop here'' as positional adjustments rather than a global stop.\\
- Recognition errors: The utterance was transcribed by a speech-to-text service, so if the utterance seems incomplete, vague, or misrecognized within this context, make a best-effort guess based on nearby words and the current trajectory to resolve any recognition errors.\\
- Irrelevant utterances: If the utterance is still irrelevant after resolving recognition errors, then ignore it if it is unlikely to be directed at you or that do not clearly include a directive about how the robot's position, velocity, or force (e.g., ``I'm sore'', ``I'm happy today'', and ``I feel fast right now'').\\
- Consider the trajectory stage based on direction:\\
\indent - Waypoints that are not near a body landmark but come before waypoints that are represent the robot moving toward the person for interaction.\\
\indent - Waypoints in contact with or near a body landmark indicate interaction.\\
\indent - Waypoints that are no longer near a body landmark but come after waypoints that were represent the robot moving away from the person following interaction.\\
- Unclear utterances: If the utterance is unclear (``This doesn't feel good'') and does not specify a target parameter (force, velocity, attract, stop) or magnitude/scope of change, do not apply a change and instead output a short and open-ended clarifying question. Do not hint at possible parameters (speed, position, force, stop) or suggest specific adjustments in your question unless the utterance clearly specifies it. If a landmark is mentioned in the utterance, then include it in your question.\\
- Always set the field `\verb|clarification: true|' if the user's utterance was unclear and requires the follow-up clarification prompt. Otherwise set `\verb|clarification: false|'.\\
- References to previous utterances: The user may make requests that refer to previous changes (e.g., ``Actually, a little more'', ``Forget what I just said'', etc.). In these cases, user the prior utterance(s) to interpret the intent:\\
\indent - If the utterance lacks a clear target but refers to the last modified quantity/location (``a little more'', ``reduce it slightly''), apply the change to that same parameter. This should essentially amplify or deamplify the previous change. Infer how strong or subtle the multiplier should be with respect to the previous value.\\
\indent - Undo: If the utterance implies an undo of the previous change, revert to the value before the last change by applying the reciprocal multiplier.\\
- ``Up'' and ``down'' position reasoning: The user may describe adjustments using directional terms such as ``up'', ``down'', ``higher'', ``lower'', etc. In these cases:\\
\indent 1. Anchor point: Identify the current body landmark from the YAML block listing the nearest landmark to each waypoint.\\
\indent 2. Target search: Reference the list of detected landmarks to select the target landmark:\\
\indent\indent - Relative ordering rule:\\
\indent\indent\indent - For limbs: move stepwise along the natural distal-to-proximal order for ``up'' and the reverse for ``down'' movement (e.g., hands to wrist to elbow to shoulder, foot to ankle to knee).\\
\indent\indent\indent - For torso/head: move from lower to upper for ``up'' and the reverse for ``down'' movement.\\
\indent\indent - The target should be the immediate anatomical landmark above or below the anchor, depending on the utterance.\\
\indent 3. Output the target landmark and update the changes accordingly. Unless the user specifies otherwise, treat relative position changes as attraction changes.\\
Example 1:\\
- History:\\
\indent Previous Utterance: ``Go further from my mouth.''\\
\indent Previous Response:\begin{verbatim}
        mouth:
            attract: 1/2.0
        \end{verbatim}
- Current utterance: ``Undo that.''\\
- Output:\begin{verbatim}
    mouth:
        attract: 2.0
    I'm coming closer to your mouth to undo the
      previous change.
    \end{verbatim}
Example 2:\\
- History:\\
\indent Previous Utterance: ``Apply less force around my knee.''\\
\indent Previous Response:\begin{verbatim}
        [left/right] knee:
            force: 1/2.0
\end{verbatim}
- Current utterance: ``Less.''\\
- Output:\begin{verbatim}
    [left/right] knee:
        force: 1/2.0
    I'm applying even less pressure to your knee.
\end{verbatim}
The following is the given YAML block:\\
\textbf{\emph{[Trajectory represented in YAML]}}\\
\\
The following is the user's utterance:\\
\textbf{\emph{[Detected user utterance]}}\\
\\
The following is the history of previous utterances and responses, if any:\\
\textbf{\emph{[Conversation history]}}\\
\\
Output your response in the form of the given YAML block with any necessary adjustment changes based on the person's utterance. Only include waypoints and fields that changed in the response YAML (so nothing with a value of 1.0). If nothing changes, output nothing. Make sure the output is wrapped in yaml, with \verb|yaml <your yaml block>|. After outputting the YAML, also output a very concise, natural-sounding, single sentence that confirms the most significant change being made to the robot's trajectory (e.g., ``I'm decreasing the pressure by half.''). Only include this sentence if a change is made. Do not explain or justify it.
}

\subsection{Dynamic Content}
There are three sections in the prompt to be filled in dynamically based on the inputs. First of all, ``Trajectory represented in YAML'' refers to the YAML representation of the input trajectory, and below we give one example of a full trajectory for the bathing task:\begin{verbatim}
waypoint 1:
    nearest landmark: none
waypoint 2:
    nearest landmark: none
waypoint 3:
    nearest landmark: left wrist
waypoint 4:
    nearest landmark: left elbow
waypoint 5:
    nearest landmark: none
waypoint 6:
    nearest landmark: none
waypoint 7:
    nearest landmark: left wrist
waypoint 8:
    nearest landmark: left elbow
waypoint 9:
    nearest landmark: none
waypoint 10:
    nearest landmark: none
waypoint 11:
    nearest landmark: left wrist
waypoint 12:
    nearest landmark: left elbow
waypoint 13:
    nearest landmark: none
waypoint 14:
    nearest landmark: none
waypoint 15:
    nearest landmark: left wrist
waypoint 16:
    nearest landmark: left elbow
waypoint 17:
    nearest landmark: none
waypoint 18:
    nearest landmark: none
waypoint 19:
    nearest landmark: none
\end{verbatim}

Next, ``Detected user utterance'' refers to the user utterance picked up from the speech-to-text service, in plain string without any formatting. This can be any of the examples shown in Figure~\ref{fig:examples} in the main text.

Finally, ``Conversation history'' refers to the most recent user utterance and YAML response from the LLM, formatted in the same style as the two examples included in the prompt itself. If there is no previous user response, the content here is simply left empty.

\section{Additional Efficacy Results}

\begin{figure}[tbh]
  \centering
 \includegraphics[width=0.6\linewidth]{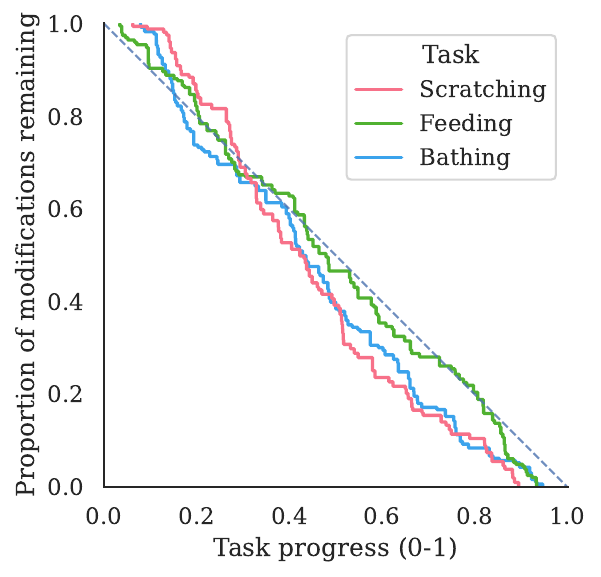}
  \caption{Empirical complementary cumulative distribution function (CCDF) showing the ratio of remaining modifications as a function of the progress of each task, from all trials with our method or unidirectional communication. Task progress is based on time, normalized to a 0--1 scale.}
  \Description{Plot showing the curves for all three tasks starts initially above the diagonal, then crosses the diagonal around 0.15--0.3, and then subsequently remaining below the diagonal.}
  \label{fig:ccdf}
\end{figure}

Figure~\ref{fig:ccdf} shows the empirical complementary cumulative distribution function (CCDF), which models the ratio of modifications that has \emph{not} been made as a function of task progress. This is computed from the timestamp of utterances for each trial involving either our method or the unidirectional communication ablation, only counting utterances directly corresponding to the goal motion aspect of each task (position for scratching, velocity for feeding, and force for bathing). The dashed diagonal line represents making modifications at a constant rate throughout task progression.

We see a common trend in all tasks where the CCDF starts above the diagonal at the beginning, crosses the diagonal, and then remain below the diagonal. This indicates that the users tend to observe the robot's behavior at the start, then make frequent modifications to reach the task objective around a task progress of 15\%--30\%, and lastly finish with less frequent modifications for small adjustments. Specifically for scratching and bathing tasks, participants typically make over 80\% of their modifications before the task progression reaches 65\%. For the feeding task, since the participants tend to increase the velocity throughout the task, trials would finish quickly after the velocity reaches a degree that the participants are satisfied with; hence we see a sharp drop in the CCDF only close to the end of the task progress (around 80\%). This tendency to make more adjustment earlier in the task progression supports the effectiveness of our trajectory modifications.

\end{document}